\documentclass[10pt, conference, compsocconf]{./IEEEtran}
\pdfoutput=1
\usepackage{comment}
\usepackage{cite}
\usepackage[pdftex]{graphicx}
\usepackage[cmex10]{amsmath}
\usepackage[tight,footnotesize]{subfigure}
\usepackage[font=footnotesize]{subfig}
\usepackage{url}
\usepackage{lipsum}
\begin{document}
\title{Learning to Grasp from 2.5D images: a Deep Reinforcement Learning Approach}
\author{
    \IEEEauthorblockN{Alessia Bertugli}
    \IEEEauthorblockA{
        University of Modena and Reggio Emilia \\
        Modena, Italy\\
        alessia.bertugli@unimore.it}
    \and
    \IEEEauthorblockN{Paolo Galeone}
    \IEEEauthorblockA{ZURU Tech \\
    Modena, Italy\\
    paolo@zuru.tech}
}

\maketitle

\begin{abstract}
In this paper, we propose a deep reinforcement learning (DRL) solution to the grasping problem using 2.5D images as the only source of information. In particular, we developed a simulated environment where a robot equipped with a vacuum gripper has the aim of reaching blocks with planar surfaces. These blocks can have different dimensions, shapes, position and orientation. Unity 3D~\footnote{https://unity3d.com} allowed us to simulate a real-world setup, where a depth camera is placed in a fixed position and the stream of images is used by our policy network to learn how to solve the task. We explored different DRL algorithms and problem configurations. The experiments demonstrated the effectiveness of the proposed DRL algorithm applied to grasp tasks guided by visual depth camera inputs. 
When using the proper policy, the proposed method estimates a robot tool configuration that reaches the object surface with negligible position and orientation errors. This is, to the best of our knowledge, the first successful attempt of using 2.5D images only as of the input of a DRL algorithm, to solve the grasping problem regressing 3D world coordinates.
\end{abstract}


\IEEEpeerreviewmaketitle
\section{INTRODUCTION}
\label{sec:Introduction}
In industrial environments, manipulator robots are usually designed to solve precise and predefined tasks. However, there are situations where it may be required to generalize the behaviour of the robots due to variations of size, shape, position, and orientation of the object to grasp. In these cases, the development of solutions according to mainstream standard computer vision and robotic control approaches can be complex and may lead to customized algorithms that cannot be easily generalized to different scenarios. Deep Reinforcement Learning addresses this task by merging the reinforcement learning and the deep learning domains, approximating the policy to learn with a deep neural network. The DRL framework has been applied to several tasks achieving impressive results. Although this success, to the best of our knowledge, no DRL algorithm has effectively solved the specific manipulation and grasping task.

The first contribution of this work is the development of a simulated grasping environment consisting of a manipulator robot equipped with a suction cup as a gripper. This environment is based on \emph{Unity 3D} together with a modified version of \emph{ML-Agents} \cite{MLAgentsPaper}, and has been used to train the proposed algorithm.
The second contribution is the DRL system that has been trained to grasp blocks characterized by random parameters in terms of dimensions, shape, position, and rotation relying only on visual data. The reinforcement learning problem is modelled so that state corresponds to the 2.5D grabbed image and the action corresponds to the manipulator tool pose. At every trial, the algorithm receives a reward in response to each action and uses the rewards to update the policy approximated by a deep neural network. Setting-up an environment using a depth-camera is easy but exploiting all the information that a 2.5D image contains is a challenging task. Solving this task can bring a lot of flexibility at a relatively low cost. We propose an analysis based on ones of the most important DRL algorithms, varying configurations of visual inputs and rewards. We used and compared three policies: Proximal Policy Optimization (PPO), Trust Region Policy Optimization (TRPO) and Deep Deterministic Policy Gradient (DDPG). Our results suggest that DRL is a promising approach to solve grasping problems that require precision.

\section{RELATED WORKS}
\label{sec:Related works}
Deep reinforcement learning has been applied to solve several tasks such as learning to play video-games and robotics problems \cite{deepQN}. In particular, it has been applied to grasp tasks with manipulator robots equipped with grippers, to locomotion tasks and also to humanoid robots \cite{loc} \cite{gibsonenv}.
These problems are currently solved in a simulated environment. Several works obtained good results, like \cite{icra2018} that involves the simulation of a \emph{Kuka} robot and its working environment to develop a solution based on deep reinforcement learning algorithms, to solve the grasping problem. 
Another recent work is \cite{handsmainpulation} that simulate four complex tasks of dexterous manipulation based on deep reinforcement learning. It uses a policy gradient method, in particular, \emph{Natural Policy Gradient} \cite{NPG} in combination with an imitation learning algorithm called \emph{Behavioral Cloning} \cite{BC} which learns a policy through supervised learning to mimic the demonstrations of an expert.

\emph{Gibson Virtual Environment} \cite{gibsonenv} is an open-source perceptual and physics simulator. This work can be seen as a bridge between learning from a simulator and transfer learning. The main goal of \emph{Gibson environment} is to facilitate transferring models trained in a simulated environment to the real world.

In particular, it ensures semantic complexity of the simulated environment that aims to be a good replica of the complex real-world and it renders visual observations that are as much as possible close to what a real camera captures. It employs a \emph{VGG-16} neural network for making rendering more like real images (forward function) and a network that makes real images look like rendering (backward function).

Only a few works ~\cite{opnairobtask, tl} achieved the complex goal of transferring the knowledge learned in a simulated environment to the real world. The core of these approaches is \emph{generalization}, obtained with randomization. In \cite{opnairobtask} authors exploit the full state observability in the simulator to learn better policies which take as input partial observation like RGB-D images. They use an \emph{actor-critic} algorithm \cite{ac1} \cite{ac2} in which the critic is trained on full states while the actor takes images as input state. Then they use domain randomization and applied transfer learning to real robot tasks such as pushing, picking and moving blocks.

The second work ~\cite{tl} uses \emph{progressive networks} to bridge the reality gap and successfully transfer policies learned in simulation to the real world. \emph{Progressive networks} aim to reuse everything from low-level visual features to high-level policies for transferring knowledge to new tasks. They use only a deep reinforcement learning approach with sparse rewards. RGB images are used as a state while joint velocity as actions.

\emph{Actor-critic} algorithm with multiple asynchronous agents is used for training in simulation with a CNN (Convolutional  Neural Network) followed by an LSTM (Long-Short Term Memory). They use discrete actions to control different degrees of freedom of the simulated robot. After training in the first simulated environment, a new network is initialized with lateral, non-linear connections to each convolutional and recurrent layer of the previously trained network. Then the new network is trained in real-world tasks, using the features learned by the first network in the simulated environment.

Although the previous works seem to obtain good results using deep reinforcement learning, there are a lot of other works that do not show these satisfactory results. A significant example is \cite{MLAgentsPaper}, which shows good results mainly on tasks with vector observations as input, while most of the tasks with visual observations perform badly.

Reward functions are difficult to design and their tuning depends on the specific problem; moreover, their design heavily affects the learning process. 
It is common that during the training process the \emph{cumulative reward} given by the policy stops growing and starts swinging around a local optimum, increasing variance around these values.

Another problem in deep reinforcement learning, and deep learning in general, is hyper-parameters tuning. Deep reinforcement learning problems involve a huge number of hyper-parameters that affect training process. It is necessary to tune hyper-parameters such as the learning rate, batch size, random seed, architecture of the policy networks, together with correct reward functions, defining and normalizing them before feeding the network. In \cite{rlmatters} the authors introduce a large comparison of state of the art problems, algorithms and implementations to highlight how deep reinforcement learning is still heavily weak to hyper-parameters variation.

\section{PROBLEM SETUP}
\label{sec:Problem setup}
 Deep reinforcement learning is a promising field to solve robotics tasks because it can solve continuous space actions problems using only raw data taken from a camera. Moreover, no manual features extraction is necessary beyond the one performed by the neural network. The goal of the project is evaluating how DRL can solve a difficult problem such as sensor-driven grasping objects with a flat gripper. As explained in section \ref{sec:Related works}, the literature shows conflicting results in deep reinforcement learning for grasp tasks. Traditional approaches based on classical computer vision are more susceptible to camera parameters variations, blocks position and rotations, light variance etc. Moreover, it is difficult to find another approach that generalizes to a random position, rotation and shape of the blocks.

\begin{figure}[t]
    \centering
	\includegraphics[width=0.8\linewidth]{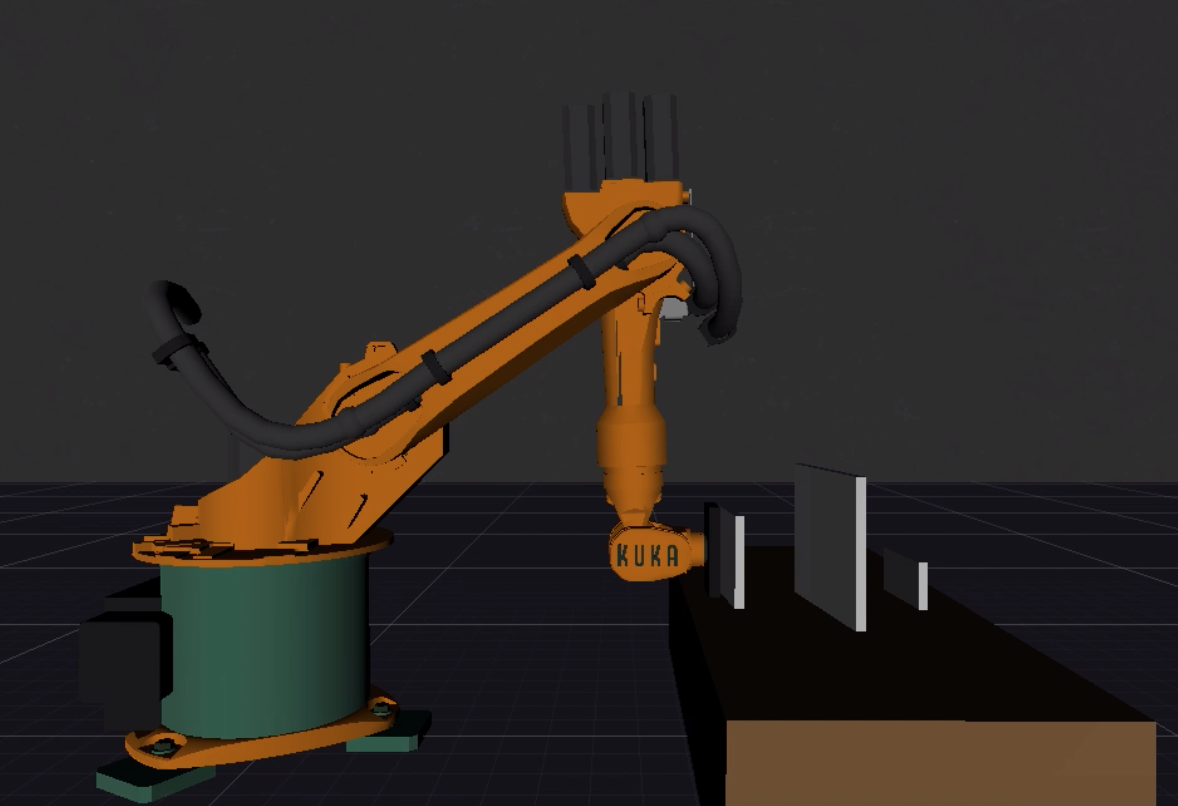}
	\caption{Simulated environment with Kuka KR16 equipped with suction cup. On a support three blocks are placed. The depth camera is placed near the base frame of the robot.}
	\label{fig:scene}
\end{figure}

The final goal is teaching to a robot, in our case a simulated Kuka KR16, to reach flat objects placed on a support in front of the robot. It has to place its end-effector in front of the first block, in a position that leads to picking the object, with the correct orientation. As far as we know, deep reinforcement learning has never been applied to suction grasping problems, which requires accurate estimation of position and orientation. The simulated environment implemented with \emph{Unity} is shown in \ref{fig:scene}.

\subsection{Depth visual observations}
\begin{figure}[b]
    \centering
	\includegraphics[width=0.8\linewidth]{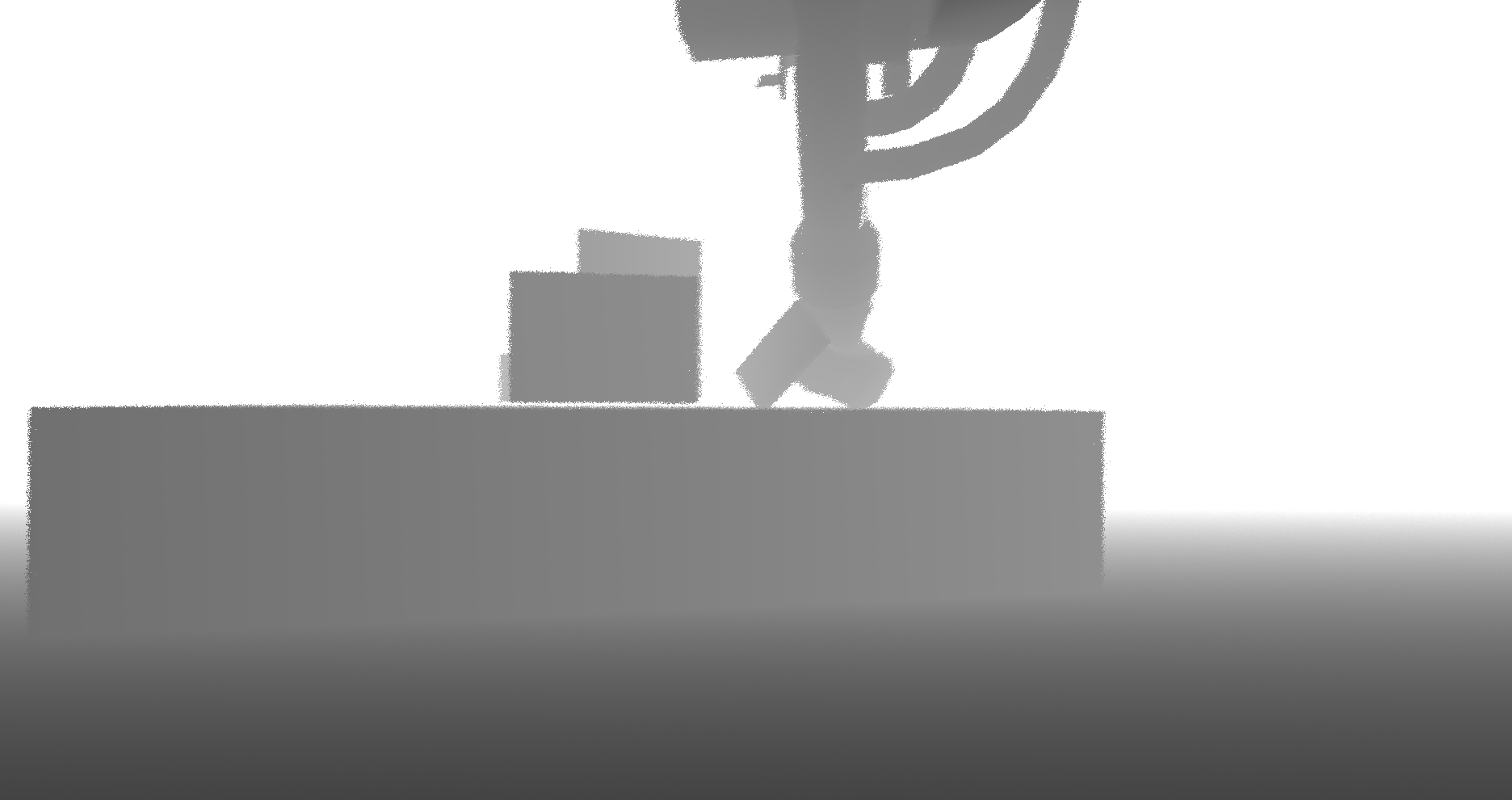}
	\caption{Depth camera view with Gaussian noise.}
	\label{fig:depth}
\end{figure}

We simulate depth camera adding Gaussian noise to make more realistic images. In \ref{fig:depth} depth camera image with Gaussian noise is shown.
ML-Agents visual encoder input is modified clamping it to the minimum and maximum range of a real camera to simulate a real depth input. In this way, when a trained model is used to make inference on real images, it does not need to make any pre-processing phase (or less as possible). For depth input we scale them between $[0.4, 2]$ meters range, try to simulate a real meters coordinates, that could generate a real depth camera. In particular, we refer to \emph{Orbbec Astra S} camera that has exactly a 0.4 to 2 meters tracking range as shown in \ref{eq:clamp}.
\begin{equation}
	new_v = min_v + actual_v \cdot \frac{max_v - min_v}{255.0}
\label{eq:clamp}
\end{equation}

We also introduce the possibility to treat differently inputs from a different type of camera: depth and RGB.

ML-Agents works with a flow of images, a stream of consecutive images that are passed through convolutional layers to extract features. Our tasks a single configuration for each episode. This makes the task solvable only looking at the first frame of the stream; there is no need for any extra information. Indeed during an episode, the agent could try different actions, but the environment configuration does not change. Moreover, the camera is positioned on the base of the robot, so images during action can show the robot that hides blocks. In such a context, frame grab can be helpful to simplify the learning process and to avoid images that show robot movements.

\subsection{Generalizations}
To make the learning process general, we applied some random transformation to the environment at the beginning of every episode. If the environment is static, the learning process risks overfitting a particular configuration (blocks position and rotation), hence, an agent trained on a static environment will be only able to grasp objects in the only configuration seen during the training phase, failing in all the others. We randomize blocks position, orientation and, scale; light position and intensity and camera position, orientation and clip planes.
\begin{figure}[ht]
	\centering
	\subfigure[Block with parallelepiped shape.]{
		\includegraphics[width=2.83cm,keepaspectratio]{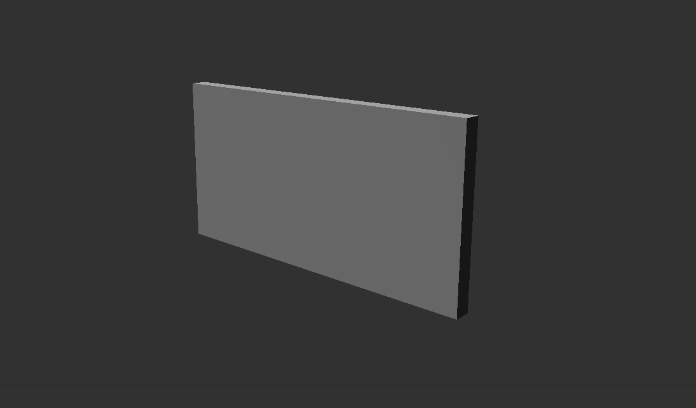}
	}
	\subfigure[Block with L shape.]{
		\includegraphics[width=2.45cm,keepaspectratio]{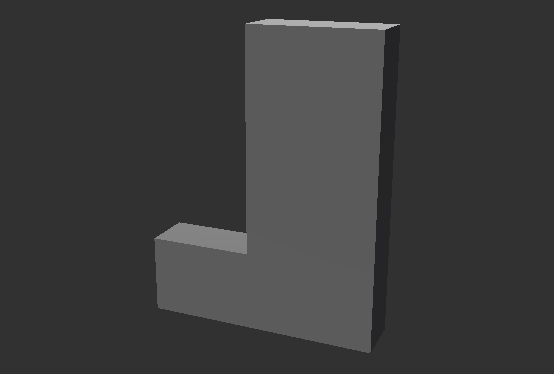}
	}
	\subfigure[Block with U shape.]{
	\includegraphics[width=2cm,keepaspectratio]{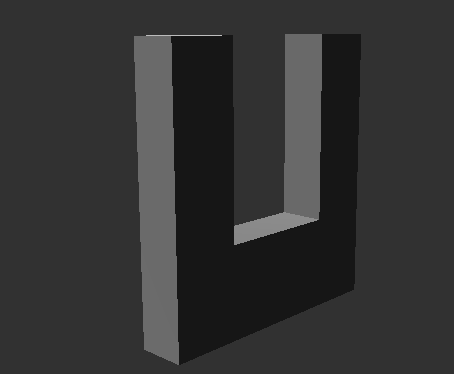}
}
\caption[Blocks with different shapes and dimensions.]{Blocks with different shapes and dimensions.
}
\end{figure}

\subsection{Rewards design}
Sparse rewards are given only when a specific event happened, for example when a goal is reached or the agent hits something in the environment. This is suitable for simple problems characterized by discrete actions (like e.g. choosing an action that leads to picking an object instead of another), substantially when actions space and solutions space are relatively small. If the problem has these properties, using sparse rewards is the best way to solve it. This is because sparse rewards are simply scalars, so it is easy to find a good reward design. However, this is not the case of many robotics tasks, especially for our problem that is characterized by huge actions space and specific target position and rotation. For this reason, dense rewards are used. Dense rewards are given at each action and give to the agent the information about how much is approaching the target. It is necessary to design functions that properly compute this measure. Rewards design is one of the challenges of reinforcement learning. With no rewards, the agent is not able to discriminate if an action is good or bad. Conversely, exceeding with rewards may lead to convergence to local minima in gradient descent. Indeed when an agent gets small positive rewards at each action, network parameters, such as \emph{beta} and \emph{learning rate}, decrease reducing exploration in favour of exploitation and accordingly stabilizing to a local minimum.

We implemented five functions to define rewards. Among them three function providing rewards only at the occurring of a specific event. In particular, a small positive reward of $0.1$ is given when the suction head touched the target. This reward is given because during the first steps, with random actions, it is really difficult for the agent reaching the correct position, so it can be useful for the learning process to give a small positive reward in response to an action that brings the end-effector to touch the target. A negative reward of $-0.1$ is given when undesired collisions happen. In particular, we consider undesired collisions, collisions between each axis of the robot and each block and support, suction head and the support. In this way, we can implement \emph{obstacle avoidance} simply giving a negative reward for each undesired collision. A positive reward is given if the action chosen takes the end-effector within a range of positions, then a further positive reward is given if also rotation coordinates are between defined values. To give a correct reward for the position we define a \emph{tooltip}, that is a 3D point position on the centre of the suction head surface. So we consider a good action an action that takes the \emph{tooltip} position coinciding with block centroid position plus an offset along \emph{z} axis that considers that centroid of the block is within the centre of the mesh. For rotation we check only rotation around \emph{y} axis. In particular we set a range of $[-0.1, 0.1]$ meters around block centroid along \emph{x} and \emph{y} axes and a range between $[0.01, 0.02]$ meters for \emph{z} axis, while for end-effector rotation we consider a range of $[-10, 10]$ degrees from block rotation. We set a reward of $0.5$ if the correct position is reached and another reward of $0.5$ if also rotation is reached. Having only these three rewards functions we may have episodes where no rewards are received. For this reason, we implement the other two functions providing progressive rewards after each action informing the agent on the effectiveness of the action chosen. The first function is shown in \ref{eq:dir_func}.
\begin{equation}
r_\text{fmt} = k_{1} \cdot f_\text{mt}
\label{eq:dir_func}
\end{equation}
where 
\begin{equation}
f_\text{mt} = \vec{v} \cdot \vec{n}
\label{eq:fmt}
\end{equation}
with $v$ that is velocity computed considering the previous position and the current one, while $\vec{n}$ is the difference between the target position and the current end-effector position, normalized. In this way, the function is positive if the action chosen makes the end-effector moving towards the target and is negative if the action chosen moves the end-effector away from the target.
A second function is implemented in \ref{eq:or_func}.
\begin{equation}
r_{fft} = - k_{2} \cdot f_\text{ft}
\label{eq:or_func}
\end{equation}
where
\begin{equation}
f_\text{ft}  = \vec{n} \cdot \vec{z_\text{ee}}
\label{eq:fft}
\end{equation}
with $z_{\text{ee}}$ which is the normal vector to the end-effector face. A positive reward is given when direction of \emph{z}-axis of end-effector frame is opposite to block \emph{z}-axis, a negative one otherwise.
$k_{1}$ and $k_{2}$ are constants empirically defined respectively as $0.03$ and $0.01$.

\subsection{Curriculum learning design}
Curriculum learning gives the possibility to increase the task difficulty during the training process. In particular, it can be useful to restrict the area that gives large rewards when the cumulative reward overcomes a predefined threshold. Each threshold defines a step of learning. We vary the area around the block that defines the best position and rotation of the end-effector. \emph{x} and \emph{y} position limits, \emph{z} position upper limit and \emph{y} rotation limits are modified at each step, restricting progressively the goal area. 19 lessons are implemented starting form a negative threshold and increasing it at each constant and durable cumulative reward increment. The last lesson provides an area of 0.01 meters variation around \emph{x}, \emph{y} and \emph{z} axes for position and 2 degrees angle around \emph{y} for rotation.

\section{REINFORCEMENT LEARNING ALGORITHMS}
\label{sec:Reinforcement learning algorithms}
Reinforcement learning is based on an autonomous agent observing a state $s_{t}$ from its environment at time $t$. The agent interacts with the environment by taking an action $a_{t}$, than the agent and the environment go to a new state $s_{t+1}$ based on the current state and the chosen action. At each transition to a new state, a single or multiple rewards are provided as scalars and then embodied in $r_{t+1}$ value given to the agent as feedback. The goal of the agent is to learn a control strategy called policy $\pi$ that maximize the expected cumulative rewards and/or maximizing a value function $V^\pi$ estimating the expected cumulative reward of a given state. Indeed what matter in reinforcement learning problems is expectations \cite{surv2}. Reinforcement learning can be described as a Markov Decision Process (MDP) which can be considered as episodic, so the state is reset after each episode with length $T$. These elements (states sequence, actions, rewards within an episode) form the trajectory of the policy, where the policy can be represented by \ref{eq:policy}.
\begin{equation}
    \pi : S \rightarrow p(\mathcal{A} = a | \mathcal{S})
	\label{eq:policy}
\end{equation}
where $\mathcal{S}$ is the set of states and $\mathcal{A}$ is the set of actions.
The policy can be considered a mapping between states and actions which maps a state to an action or a distribution over actions and its goal is finding an optimal mapping between them. Deep reinforcement learning consists of optimizing a policy using a deep neural network as an approximation. Optimal criteria represent how good the model fit data and this measure in DRL is \emph{future rewards}. There are two optimal criteria to maximize future rewards: \emph{value function}, to evaluate the state on base of the probability of future rewards and \emph{policy}, to guide an agent in the choice of an action given the state. These criteria generate three classes of DRL algorithms: value-based, policy-based and actor-critic. In policy-based methods, policy outputs parameters for a probability distribution, usually modelled by Gaussian distributions for continuous actions. Policy parameters are updated to maximize expected return using gradient-based or gradient-free optimization. Policy gradients methods are often used in \emph{actor-critic} implementations. Actor-critic methods combine value function methods with policy-based methods, reducing variance introduced by policy gradient algorithms and at the same time bias introduced from value function methods. Indeed actor-critic uses value function learned to optimize policy. Our focus is on \emph{actor-critic} algorithms. In particular, three of the most important and recent ones are used: \emph{Proximal Policy Optimization}, \emph{Trust Region Policy Optimization}and \emph{Deep Deterministic Policy Gradient}. We also used PPO combining with \emph{Curriculum Learning}.

\subsection{Deep Deterministic Policy Gradient}
Deep Deterministic Policy Gradient is a model-free, off-policy algorithm using deep function approximator that can learn policies in high-dimensional and continuous action space. It is based on a policy gradient algorithm that uses a stochastic behaviour policy to improve exploration. Actions are chosen to base on the stochastic policy, to estimate a deterministic target policy which is easier to learn. DDPG uses experience replay that stores a replay buffer with the experiences of the agent during training, and then randomly sample experiences to break up the temporal correlations within different training episodes. DDPG can learn using low-dimensional observations and directly from pixels. In particular, we use \cite{ac2} version that adds normal noise $N$ to exploration actor policy $\mu'$ sampled from a noise process as shown in \ref{eq:noise}.
\begin{equation}
	\mu'(s_{t}) = \mu(s_{t} | \theta^\mu_{t}) + N
	\label{eq:noise}
\end{equation}
\newline

\subsection{Trust Region Policy Optimization}
Trust Region Policy Optimization aims to minimize an objective function guaranteeing policy improvement with non-trivial step size. Then a series of approximations are done to apply a theoretical algorithm to practice \cite{TRPO}. In TRPO an objective function is maximized subject to a constraint on the size of the policy update. It is formally written in \ref{eq:trpo}.
\begin{equation}
max_{\theta} \hat{E}_{t} \bigg [\frac{\pi_{\theta}(a_{t} | s_{t})}{\pi_{\theta_{old}}(a_{t} | s_{t})} \hat{A}_{t}\bigg]
\label{eq:trpo}
\end{equation}
subject to \ref{eq:trpo2}
\begin{equation}
\hat{E}_{t}[KL[\pi_{\theta_{old}}( \cdot | s_{t}), \pi_{\theta}( \cdot | s_{t})]] \leq \delta
\label{eq:trpo2}
\end{equation}
where KL is Kullback - Leibler divergence that is a measure of how one probability distribution diverges from a second expected probability distribution.
A constraint can be added in the form of penalty solving the unconstrained optimization problem.
\begin{equation}
max_{\theta} \hat{E}_{t} \bigg [\frac{\pi_{\theta}(a_{t} | s_{t})}{\pi_{\theta_{old}}(a_{t} | s_{t})} \hat{A}_{t}\bigg] - \beta \hat{E}_{t}[KL[\pi_{\theta_{old}}( \cdot | s_{t}), \pi_{\theta}( \cdot | s_{t})]] 
\end{equation}
In this way, for \emph{Lagrange multiplier} method optimal point of $\delta$-constrained point problem is also an optimal point of $\beta$-penalized problem for some $\beta$ and in practice, $\delta$ is easier to tune.
Using max KL instead of mean KL lead to a pessimistic bound on policy performance. TRPO has several limitations: it is hard to use with architecture with multiple outputs and empirically performs poorly with deep CNNs and RNNs (Recurrent Neural Networks). Moreover constrained gradient computation makes implementation core complicated.

\subsection{Proximal Policy Optimization}
Proximal Policy Optimization is an algorithm that reaches data efficiency and performance of TRPO, without its limitations, using only first-order optimization. It uses clipped probability ratios which form a lower bound of policy performance. It is based on an alternation between sampling data from the policy and performing subsequent epochs of optimization on the same data. Formally it can be explained introducing the probability ratio $r_{t}(\theta)$ so that $r_{t}(\theta_{\text{old}}) = 1$. PPO modify the objective to penalize change that moves $r_{t}(\theta)$ away from 1. In particular, $L^{CLI}(\theta)$, lower bound on the expected improvement of consecutive policy iteration, can be written as in \ref{eq:ppo}.
\begin{equation}
L^{CLI}(\theta) = \hat{E}_{t} \bigg [\frac{\pi_{\theta}(a_{t} | s_{t})}{\pi_{\theta_{old}}(a_{t} | s_{t})} \hat{A}_{t}\bigg] = \hat{E}_{t} [r_{t}(\theta) \hat{A}_{t}]
\label{eq:ppo}
\end{equation}
This equation has been modified as by adding a clip normalization between $[1- \epsilon, 1 + \epsilon]$ as in \ref{eq:ppo2}.
\begin{equation}
L^{CLIP}(\theta) =  \hat{E}_{t} [min(r_{t}(\theta) \hat{A}_{t}, clip(r_{t}(\theta), 1 - \epsilon, 1 + \epsilon)\hat{A}_{t})]
\label{eq:ppo2}
\end{equation}
$L^{CLIP}(\theta)$ is a lower bound on $L^{CLI}(\theta)$ with a penalty for having too large policy updates.
Another way is to use a penalty on KL divergence to achieve some target value at each policy update. Using a neural network as an approximation that shares parameters between the policy and the value function, a loss function has to be used to combine policy surrogate and a value function error term. It is done adding an entropy bonus to ensure exploration \cite{PPO}. This policy is better then TRPO in continuous control and compatible with multiple output networks, such as CNNs and RNNs.

\subsection{Curriculum learning}
Curriculum learning can be seen as a general strategy for the global optimization of non-convex functions. The basic idea of curriculum learning is to augment problem difficulty during training. At an abstract level, a curriculum can also be seen as a sequence of training criteria. Each training criterion in the sequence is associated with a different set of weights on the training examples, or more generally, on a re-weighting of the training distribution. Initially, simpler examples that can be learned easily are presented to the agent. Then the next steps involve a slight change in the weighting of examples that increase the probability of sampling slightly more difficult examples. At the end of the sequence, the re-weighting of the examples is uniform and we train on the target training set or the target training distribution \cite{CurrLearning}. In deep reinforcement learning \emph{curriculum learning} can be implemented modifying the environment to increase the problem difficulty during training.

\section{EXPERIMENTS AND RESULTS}
In this section, the experimental validation and parameters tuning process are presented. Even if an environment is well defined, or the reset logic is coherent to the problem and the reward design is done, the training process can lead to bad results. Then, a hard process of parameters tuning has to be performed to obtain the best configuration for the sensor-based grasping task. Final results of the work are shown, in particular, the results varying according to visual inputs dimensions and architectures. CNNs with a different number of convolutional layers and input images dimensions are tested. The same task is performed, in a scenario with two cameras. An additional RGB camera is fixed on the top of the support, to observe the objects. Also, frame grab working is tested. Eventually, an algorithms comparison is done using the most famous state of the art reinforcement learning algorithms. The task is addressed with DDPG and TRPO policies, with PPO plus curriculum learning and PPO with the addition of LSTM besides the CNN. For DDPG and TRPO \emph{stable baselines} \cite{stable_baselines} implementation is used.

Not all tests have been done keeping the same maximum number of training steps. The tests were interrupted when the result seemed satisfactory, or when the increase in the cumulative reward did not seem significant for a large number of steps. In the latter case, it is assumed that to obtain positive results, we should expect a much longer time (not by the purpose of this analysis) or have greater computing resources.

All the experiments have been performed using a Intel\textsuperscript{\textregistered} Core\textsuperscript{\texttrademark} i7-7820X CPU and NVIDIA\textsuperscript{\textregistered} GTX 1060.
\subsection{Sparse rewards}
To validate the analysis we tried to solve the task of taking with sparse rewards. In this test is returned only one positive reward when the action chosen by the network leads the agent to reach the optimal area. No other rewards have been prepared. The results of subsequent tests show that this type of reward is not sufficient to solve the proposed problem. The cumulative reward remains for many steps around zero with peaks in correspondence with so-called winning actions. This does not mean that the problem can not be solved with a single positive reward, but that probably takes many more training steps compared to the configuration of the problem with dense rewards. Observing the results, it can be seen that at the same number of training steps, in the case of sparse rewards the curve of the cumulative reward remains flat on zero with peaks, while in the case of dense rewards the curve is already growing.

\subsection{Visual inputs and CNN architectures}
Different input dimensions and convolutional architecture are tested with PPO. In particular we test four image dimensions with 4 layers configurations: 32x32 with 2 convolutional layers (the first one with 4 filters, kernel size [2, 2], strides [4, 4], the second one with 8 filters, kernel size [1, 1], strides [2, 2]), 80x80 with 2 convolutional layers (the first one with 16 filters, kernel size [8, 8], strides [4, 4], the second one with 32 filters, kernel size [4, 4], strides [2, 2]), 128x128 with 3 convolutional layers (the first and the second ones with 80x80 parameters with a third additional layer with 64 filters kernel size [2, 2], strides [1, 1]) and 256x256 with 4 convolutional layers (the first, the second and the third ones with 128x128 parameters with a fourth additional layer with 72 filters kernel size [2, 2], strides [1, 1]). All CNN architectures give satisfactory results in terms of cumulative reward, except for 32x32 test that led to an unstable training process.

Other visual inputs modifications are tested: using two cameras and frame grab. The using of both RGB and depth camera has resulted in a performing learning process. Cumulative reward increases during training and the agent can avoid obstacles and to reach the desired block.

\begin{figure}[t]
    \centering
	\includegraphics[width=0.8\linewidth]{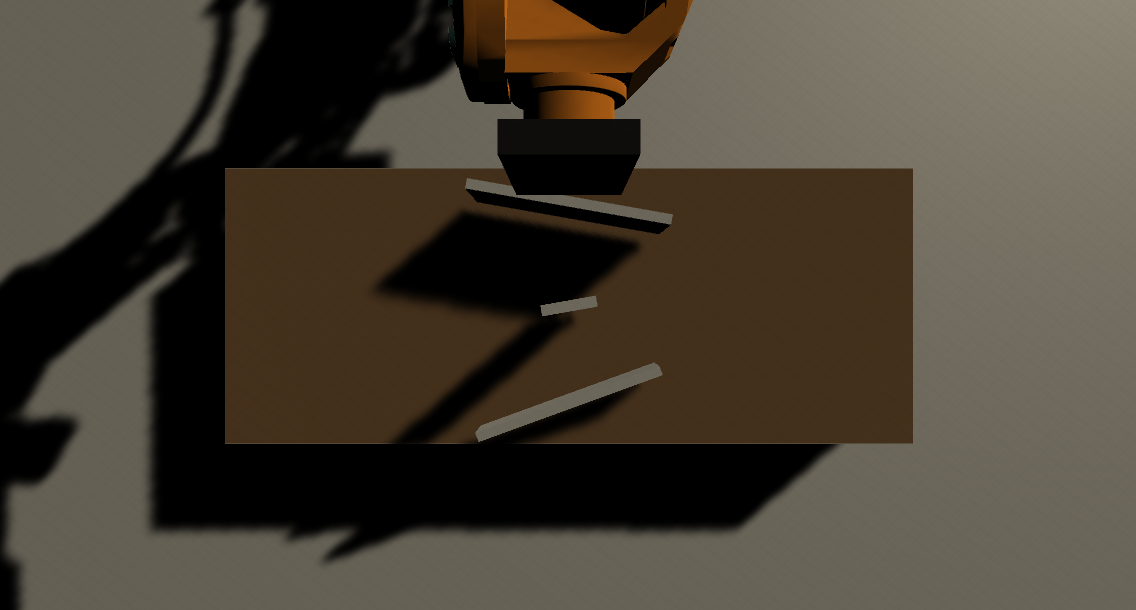}
	\caption{View from RGB camera placed over the support looking at blocks.}
	\label{fig:scene_top}
\end{figure}
Also frame grab gives results comparable to the normal process.

\subsection{Policies comparison}
Most famous reinforcement learning algorithms described in Section \ref{sec:Reinforcement learning algorithms} are compared.
DDPG performs poorly with our environment. This is probably due to the huge number of hyper-parameters DDPG requires to tune. Probably applying grid or random search on all of them better results could be reached.
TRPO gives some positive results. Cumulative reward increase overcoming 0, but it results in an unstable learning process.
PPO with curriculum learning gives the best results. The agent can reach the correct position and rotation with a little margin of error. The trend seems to be unstable, but this is due to the lesson changing after a defined value of the reward is reached.
PPO with a different network architecture is also tested: adding an LSTM besides the CNN. It does not give good results, probably because LSTMs are not particularity adapt to solve this kind of tasks. It results in a cumulative reward that oscillates around $0$ for most of the training steps, with some peaks.

\begin{figure*}[ht]
	\centering
	\subfigure[Cumulative reward of sparse rewards test.]{
	\includegraphics[width=5.2cm,keepaspectratio]{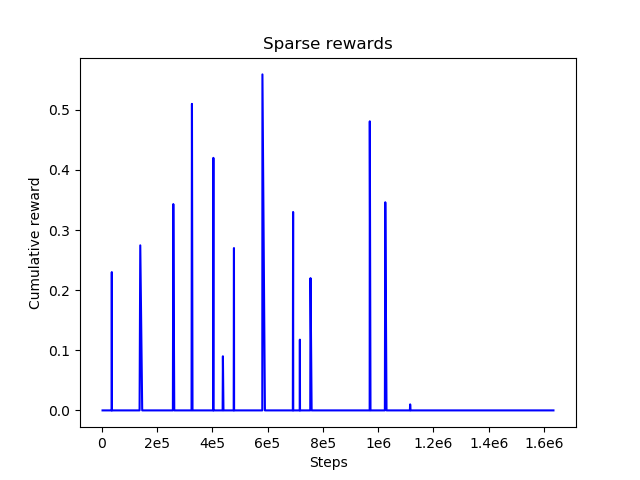}
	}
	\subfigure[Cumulative reward of test with two cameras: depth plus RGB.]{
		\includegraphics[width=5.2cm,keepaspectratio]{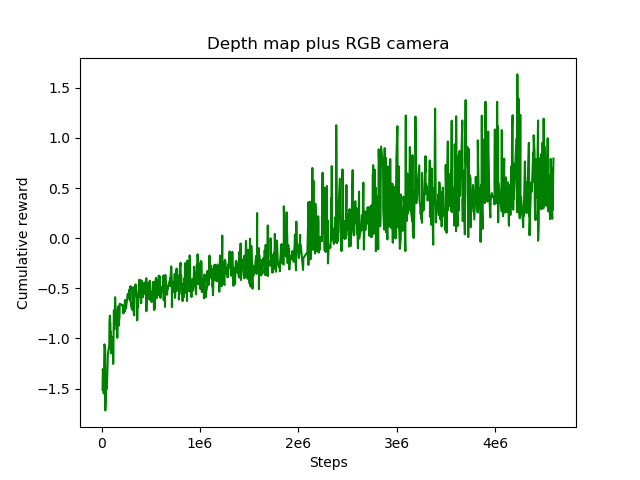}
	}
	\subfigure[Cumulative reward frame grab test.]{
		\includegraphics[width=5.2cm,keepaspectratio]{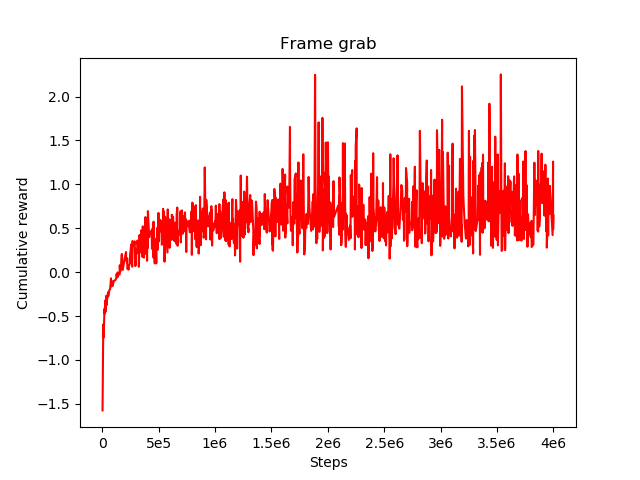}
	}
		\subfigure[Cumulative reward of 32x32 images test.]{
		\includegraphics[width=5.2cm,keepaspectratio]{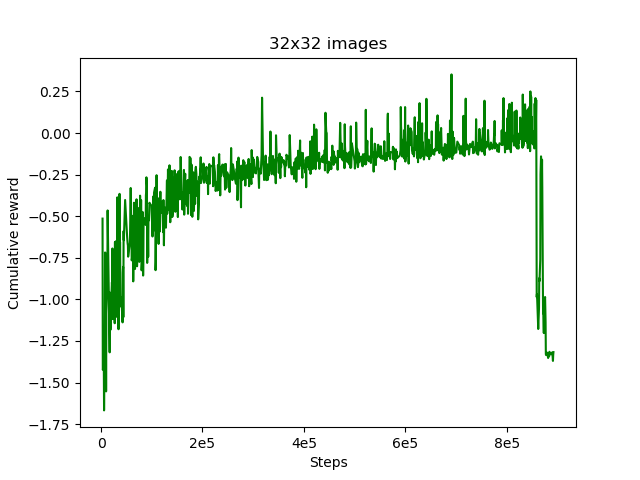}
	}
	\subfigure[Cumulative reward of 80x80 images test.]{
		\includegraphics[width=5.2cm,keepaspectratio]{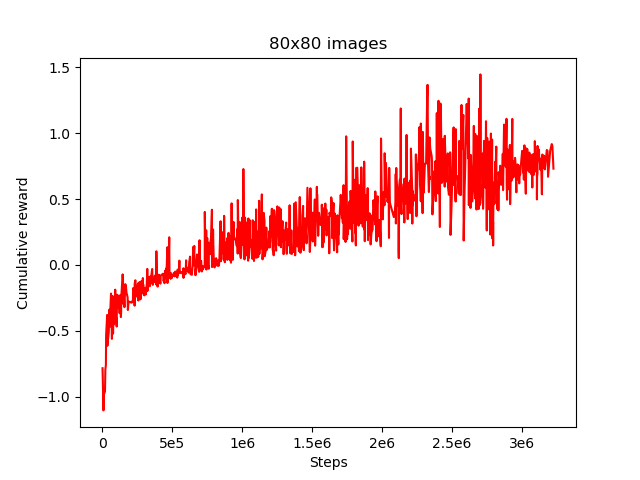}
	}
	\subfigure[Cumulative reward of 128x128 images test.]{
		\includegraphics[width=5.2cm,keepaspectratio]{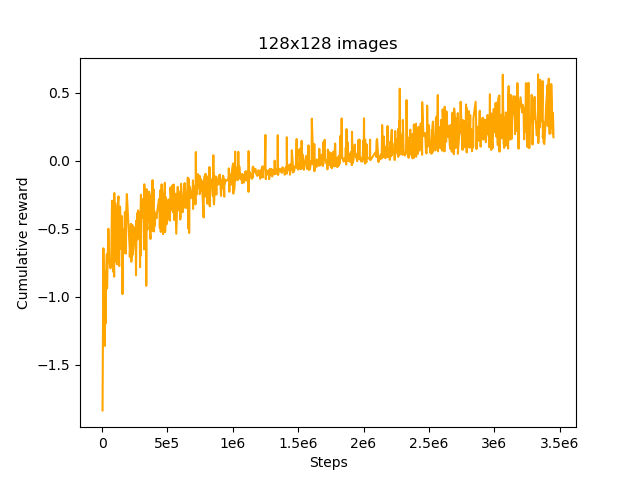}
	}
    \subfigure[Cumulative reward of 256x256 images test.]{
	\includegraphics[width=5.2cm,keepaspectratio]{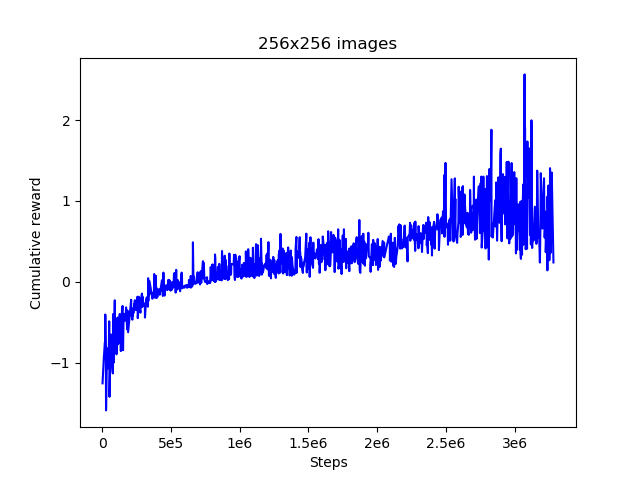}
    }
	\subfigure[Cumulative reward of PPO plus Curriculum learning test.]{
		\includegraphics[width=5.2cm,keepaspectratio]{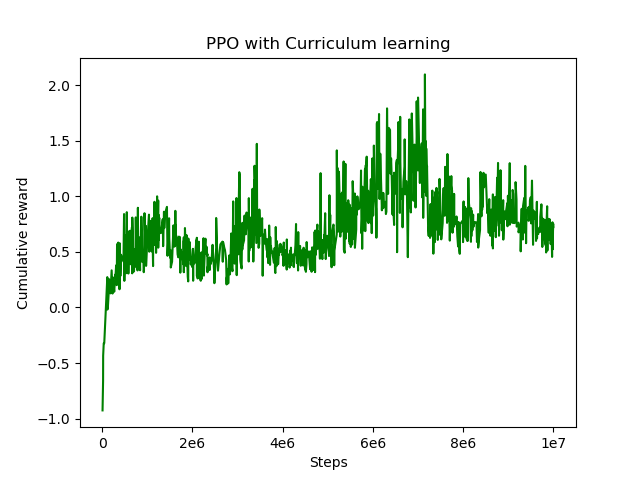}
	}
	\subfigure[Cumulative reward of PPO with CNN following by LSTM test.]{
	\includegraphics[width=5.2cm,keepaspectratio]{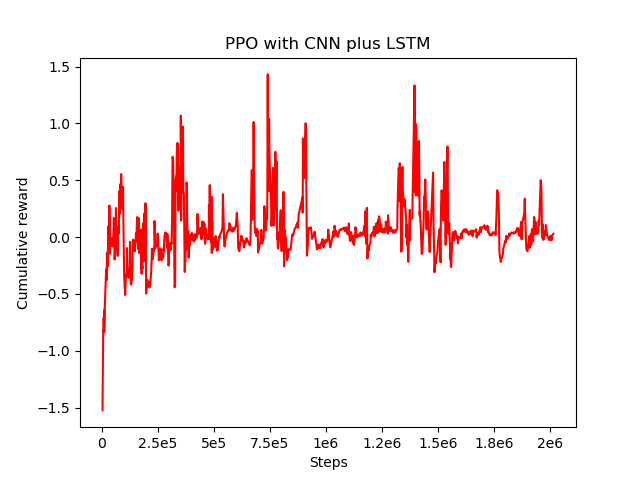}
    }
	\subfigure[Cumulative reward of TRPO test.]{
		\includegraphics[width=5.2cm,keepaspectratio]{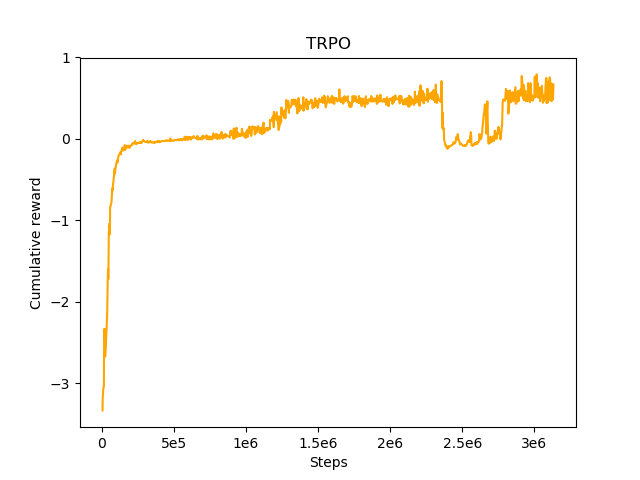}
	}
    \subfigure[Cumulative reward of DDPG test.]{
	\includegraphics[width=5.2cm,keepaspectratio]{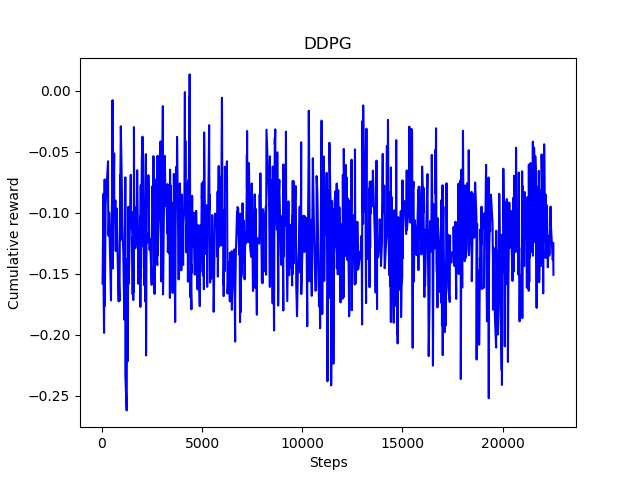}
    }
\caption[Comparison.]{Cumulative reward of all tests. %
}
\label{fig:comp}
\end{figure*}

\subsection{Hyper-parameters tuning}
Hyper-parameters tuning process is hard in all machine learning tasks and, in particular for deep reinforcement learning ones because of the huge number of parameters related both to the policy and to the CNN. Moreover, there are several factors related to task design that can affect performances such as actions space, reward values etc.. According to our analysis, parameters that mostly affect the training process are \emph{learning rate} and \emph{batch size}. For the learning rate, two decay types are analyzed (polynomial decay and exponential decay) and different starting values for both the actor and the critic. Depending on the task and configuration, the initial learning rate is set between $10^{-4}$ and $10^{-3}$. Despite batch size is usually set to a high value for continuous actions space, our experiments show that a small value (32) can bring to satisfactory results, speeding up the learning process. For PPO training \emph{beta} decay is slowed down to maintain a correct balancing between exploration and exploitation for the process.

\section{DISCUSSION AND FUTURE WORK}
With this work, we demonstrate the validity of a deep reinforcement learning approach to grasping problems with a vacuum gripper and 2.5D visual inputs.
We applied the state of the art reinforcement learning algorithms to this task and compared them.

Different types of visual inputs have been tested together with the definition of several convolutional architectures to find the best configuration. We experimented how the DRL algorithms behave with different inputs such as a single 2.5D input coming from a simulated depth-camera and two visual inputs together as input to the network using a simulated depth map and an RGB camera.

Our approach to the environment definition and reward design gives good results and this suggests that it could be applied to a real-world scenario transferring the knowledge from the simulated environment to the real one, applying some transfer learning techniques. Further research directions can be the definition of a more complete environment that simulates a point cloud together with an extension to ML-Agents that allows working with this kind of data.

\section*{ACKNOWLEDGMENT}
The authors would like to thank Dario Lodi Rizzini of the University of Parma, for several fruitful discussions and valuable advice.

\bibliographystyle{IEEEtran}
\bibliography{IEEEtranBST/IEEEabrv,IEEEtranBST/mybibfile}

\end{document}